\documentclass[letterpaper, 10 pt, conference]{ieeeconf}  %

\IEEEoverridecommandlockouts                              %
                                                          
\usepackage{hyperref}
\usepackage{cite}
\usepackage{amsmath,amssymb,amsfonts}
\usepackage{algorithmic}
\usepackage{amsmath}
\usepackage{graphicx}
\usepackage{textcomp}
\usepackage{xcolor}
\usepackage{float}
\usepackage{leftidx}
\usepackage[caption = false]{subfig}
\usepackage{graphicx}
\usepackage{balance}

\begin{document}

\title{\textbf{Tactile Aware Dynamic Obstacle Avoidance in Crowded Environment with Deep Reinforcement Learning}}

\author{Yung Chuen Ng\textsuperscript{1}, Qi Wen (Shervina) Lim\textsuperscript{1}, Chun Ye Tan\textsuperscript{2}, Zhen Hao Gan\textsuperscript{1}, Meng Yee (Michael) Chuah\textsuperscript{1} \thanks{$^{1}$Institute for Infocomm Research (I\textsuperscript{2}R), A*STAR, Singapore, {\tt\small {ng$\_$yung$\_$chuen, shervina$\_$lim, gan$\_$zhen$\_$hao, michael$\_$chuah}@i2r.a-star.edu.sg}} \thanks{$^{2}$Nanyang Technological University (NTU), Singapore, {\tt\small ctan173@e.ntu.edu.sg}} }

\maketitle

\begin{abstract}
Mobile robots operating in crowded environments require the ability to navigate among humans and surrounding obstacles efficiently while adhering to safety standards and socially compliant mannerisms. This scale of the robot navigation problem may be classified as both a local path planning and trajectory optimization problem. This work presents an array of force sensors that act as a tactile layer to complement the use of a LiDAR %
for the purpose of inducing awareness of contact with any surrounding objects within immediate vicinity of a mobile robot undetected by LiDARs. By incorporating the tactile layer, the robot can take more risks in its movements and possibly go right up to an obstacle or wall, and gently squeeze past it.
In addition, we built up a simulation platform via Pybullet which integrates Robot Operating System (ROS) and reinforcement learning (RL) together. A touch-aware neural network model was trained on it to create an RL-based local path planner for dynamic obstacle avoidance. 
Our proposed method was demonstrated successfully on an omni-directional mobile robot who was able to navigate in a crowded environment with high agility and versatility in movement, while not being overly sensitive to nearby obstacles-not-in-contact. 
\end{abstract}

\section{Introduction}

Since the advent of mobile robotics, autonomous mobile robots have been increasingly deployed in busy human settings. Businesses have adopted robotics technology to streamline operations and logistics via various applications, for instance, delivery, cleaning, inspection, patrol and surveillance. However, these are typically done within structured environments such as in factories and warehouses, as dynamic obstacles still pose a challenge in robust path planning. 

To enable the use of mobile robots in human environments, it is necessary for them to safely, efficiently and robustly navigate among people moving around them. The problem of obstacle avoidance arises when a robot attempts path planning to generate a collision-free motion trajectory. In the presence of other moving obstacles, there exists an element of uncertainty where the obstacle may cross the generated motion trajectory and potentially result in collision with the robot. In normal everyday human interactions, people would predict and determine the motions of others based on observed social cues. For a robot operating in a human environment, it becomes all the more important for it to develop an ‘awareness’ to predict surrounding motions and preempt sudden movements so that path planning will be safer without accidents. This requires robots that are able to respond within human reaction time to do obstacle avoidance in a continually changing, dynamic environment and have an understanding of human movement and intentions. Take a busy hospital setting for example, a robot is required to deliver items such as medicine to patients while navigating among moving crowd of people and nurses. Therefore, there is a need to develop a dynamic and robust solution to tackle navigation in a crowded environment. 

\begin{figure}[tb]
\centering
\includegraphics[width=0.48\textwidth]{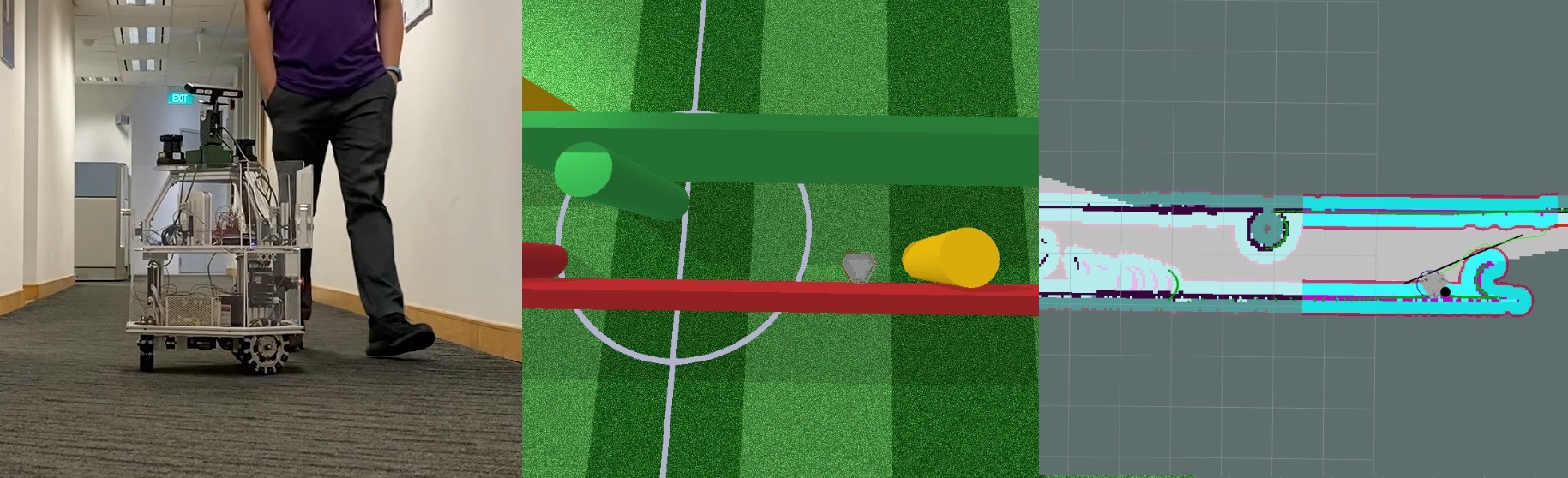}
\vspace{-4mm}
\caption{Omnidirectional mobile robot navigating in both simulation and real world. Left: real omni-robot used in this paper; Middle: Pybullet simulation setup; Right: ROS Rviz monitoring window.}
\label{fig:first_figure}
\vspace{-3mm}
\end{figure}

This paper presents a tactile layer to complement LiDAR in online local path planning for dynamic obstacle avoidance in crowded environments. to achieve this, an open-source framework is introduced to increase the sim-to-real transferability of the developed solution. In summary, the main contributions are:
\begin{itemize}
\item a tactile aware intelligence layer that informs the mobile robot of any contact in a 360\textdegree\ zone around the robot as well as the magnitude of the contact force.
\item an open-source reinforcement learning (RL) framework integrating Robot Operating System (ROS), OpenAI Gym \cite{brockman2016openaigym}, and Pybullet Gymperium \cite{benelot2018} that trains a ROS robot in Pybullet simulation for improved sim-to-real performance \cite{coumans2019}.
\item an end-to-end RL-based local path planner leveraging contact and LiDAR sensors.
\item demonstrating dynamic obstacle avoidance on an omnidirectional mobile robot in a crowded environment.
\end{itemize}

\section{Related Work}

\subsection{Omnidirectional Mobile Robot Navigation}

As an omnidirectional robot is a holonomic robot that is able to translate and rotate simultaneously, it is a popular research platform for mobile robot navigation. Previous research has applied swarm optimization \cite{swarm_optimization}, line trajectory adaption \cite{line_trajectory_adaptation} and moving obstacle-avoidance algorithm \cite{dynamic_obstacle_avoidance_omni} on omnidirectional robots to perform obstacle avoidance. These methods are not robust in avoiding dynamic obstacles in certain situations such as when obstacles approach the robot at high speed \cite{dynamic_obstacle_avoidance_omni}, sometimes leading to inevitable collisions. 

\subsection{Classical Local Path Planning Approaches}

Common classical approaches to dynamic obstacle avoidance implements straightforward kinematics and a simple algorithm for collision avoidance solely based on real-time LiDAR data \cite{dwa, apf, mcguire2018comparativebug}. However, even maps produced by common ROS-based SLAM algorithms may have errors up to 51.67 cm when compared to the ground truth \cite{lidar2d}. Another issue is that some LiDARs have near-field detection blind spots, and these issues make it difficult for the robot to determine how close it can safely get to the obstacle before it collides with it.

For these reasons, existing solutions would chose to inflate obstacles as a safety buffer to avoid having the robot move too close \cite{inflation-ros, inflation-humanoid} . However, this would make the robot take overly safe maneuvers and degrade performance in dynamic obstacle avoidance in a crowded setting. Other approaches considered the modeling and prediction of pedestrian behaviours via trajectory data obtained by external camera vision to output movements among busy crowd but these methods have heavy reliance over a map with overhead view and complete positional information of pedestrians around the subject and cannot be done with just on-board sensors \cite{vemula2017modelingcoopnav}.

In other implementations of path planning and obstacle avoidance, a robot agent is presumed to be operating in known environments  with information on obstacle positions and targets, allowing smooth simulation and run of local path planning algorithms using a global path planner. Some examples of which includes the A* and D* algorithm \cite{Stentz1995TheFD*}, triangular-cell-based map \cite{triangular_map}, and more recently the Collision Avoidance using Deep Reinforcement Learning (CADRL) network \cite{everett2018motion_cadrl}. However, in an unknown environment, the robot will only have partial information of its surroundings and must plan a path based on local sensor information for avoiding collision.

In short, there is a lack of existing works that addresses online navigation in a crowded unknown environment using only the robot's on-board sensors - a difficult problem considering the volatility in environment and the stochasticity in people's behaviour during locomotion \cite{sociallycompliant_irl, Kim2016Socially_adaptive_irl}.

\subsection{Heuristic Local Path Planning Approaches}

Researchers are increasingly drawing on and implementing artificial intelligence in robotics to do obstacle avoidance. %
Different sensors such as RGB monocular vision \cite{xie2017monocular} and LiDAR \cite{ddqn} have been tested as inputs for training a neural network. A variey of different RL algorithms such as A3C \cite{crowded_drl_2020}, D3QN \cite{xie2017monocular}, ADDPG \cite{addpg} and DDQN \cite{ddqn} have also been used. The RGB monocular vision-based obstacle avoidance method has only been tested for short term navigation up to 20 seconds \cite{xie2017monocular} while the use of ADPPG learning based mapless motion planner has not been verified with localisation in an actual navigation scenario \cite{addpg}. CADRL presents good performance in avoiding dynamic obstacles but the algorithm only allows turning in one direction \cite{everett2018motion_cadrl}. CADRL will also experience collision if the algorithm fails to detect nearby unseen and unmapped static obstacles, which is a major disadvantage.

\subsection{Deep Reinforcement Learning Frameworks}

Current solutions present RL frameworks such as \cite{benelot2018, brockman2016openaigym, openairos}. The main drawback of these frameworks is the lack of transferability from simulation to real-life environment despite potentially good experimental outcomes - fast training time, convergence, and optimal validation results. Our approach seeks to address this issue with the integration of ROS, and demonstrate this by directly applying a simulation trained RL agent onto a physical mobile robot to perform dynamic obstacle avoidance in the real-world.

\begin{figure}[tbp]
\centering
\includegraphics[width=0.4\textwidth]{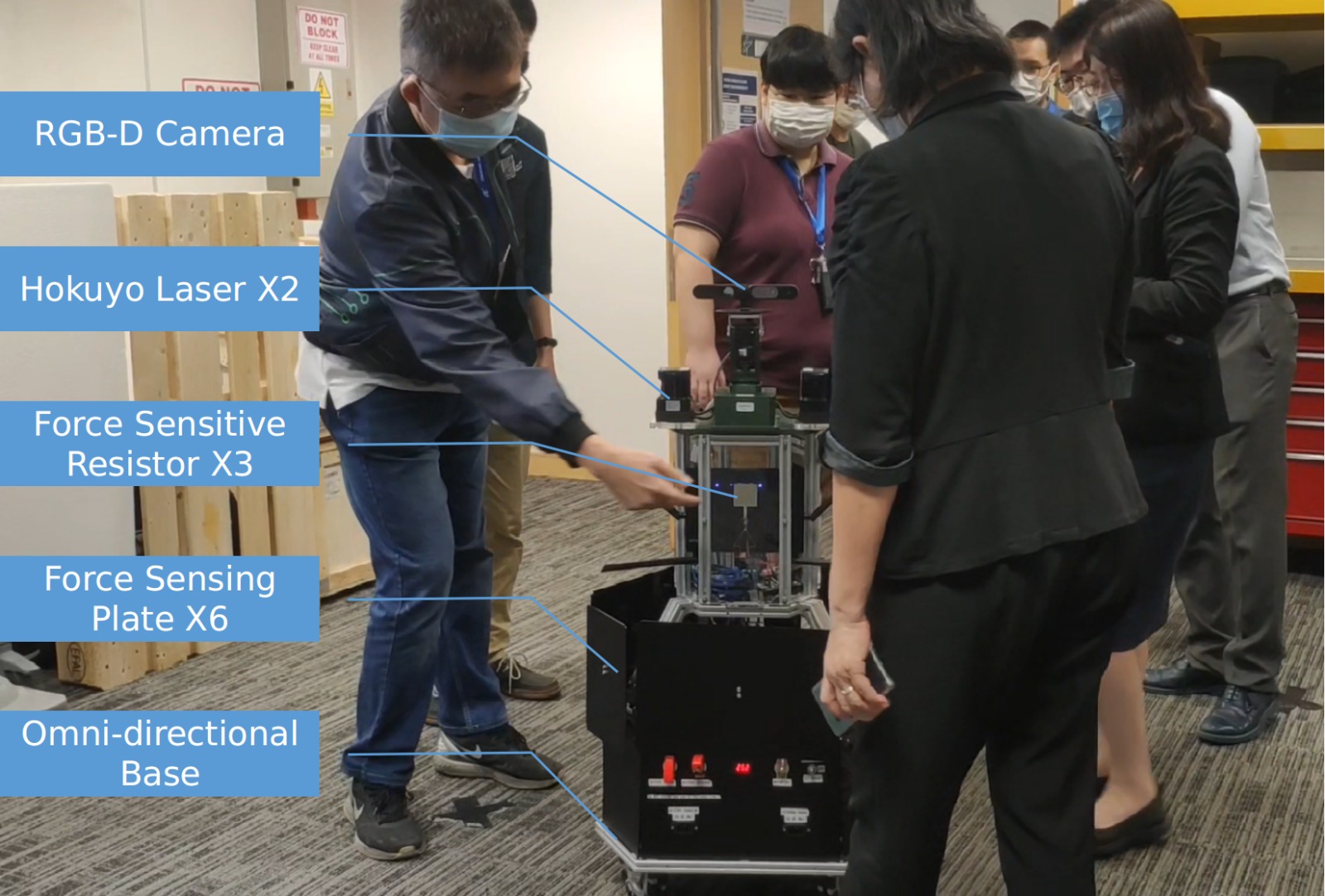}
\vspace{-1mm}
\caption{The hardware system of our omnidirectional mobile robot platform. The robot is retrofitted with the sensors: RGB-D Camera, 2 HOKUYO LiDARs, 3 thin film force sensing resistors, 6 force sensing plates. The RGB-D camera on the robot is not used in this paper, but in ongoing work.}
\label{fig:robot}
\vspace{-3mm}
\end{figure}

\section{Approach}

\subsection{Problem Definition}

Dynamic obstacle avoidance may be formulated as a trajectory optimisation problem, the long-term cost over continuous trajectories using a standard additive-cost optimal control objective is given by:

\begin{equation}
J_u(x_0)=L_f(x(t_f)) + \int\limits_{t_0}^{t_f} L(x(t),u(t)) dt \label{eq:optimization}
\end{equation}

where $J_u(x_0)$ is the continuous-time cost, $L_f(x(t_f))$ is the endpoint cost and $L(x(t),u(t))$ is the Lagrangian. 
The aim of the problem is to minimise $J_{u}$ related to the path taken by a robot traveling from a start point to an end goal. 

This paper aims to address the local path planning navigation problem via an RL-based local path planner for an omnidirectional mobile robot platform. The use of RL to address dynamic obstacle avoidance remodels the problem via a Partially Observable Markov Decision Process (POMDP) framework. The POMDP may be described as a seven-tuple:
\begin{equation}
P=f(S, A, T, R, \Omega, O, \gamma)\label{eq:fn_of_states}
\end{equation}
with state space $S$, action space $A$, state transition probability matrix $T$, reward function $R$,  the measurement from the raw sensor observation $\Omega$, observation probabilities O and discount factor $\gamma   \in   [0,1]$. 
The aim of the problem is to train an optimal policy $\pi$ that maximizes the Q-value function for every state s as a subset of S and action a as a subset of A as illustrated in the following:
\begin{equation}
Q^{\pi}(s,a) = r_{t} + \sum_{t=0}^{t-1} (\gamma^{t} r_{t}| s_{t}, a_{t})\label{eq:Q_value}
\end{equation}
with $r_{t}$ being a scalar reward received when transitioning from a state $s_{t}$ at time step t to the next state $s_{t+1}$ at time step t+1. The optimal policy that maximises Q is $\pi^{*}(s) = \arg \max_{\pi} Q(s,a)$.
This policy directly maps the state to the action. For an effective motion planner, the control frequency of 10 Hz must be guaranteed so that the robot reacts to new observations as fast as a human \cite{VANRULLEN2012995_10Hz}.

\subsection{The Proposed RL Framework}

    \textit{ros\_pybullet\_rl2} is an integrated ROS-RL framework based on the open-source \textit{rl-baselines3-zoo} \cite{rl-zoo3}, designed to train the dynamic obstacle avoidance behaviour on an omnidirectional mobile robot platform in Pybullet physics simulator \cite{coumans2019}. This framework inherits from ROS, Pybullet Gymperium \cite{benelot2018}, and Stable Baselines3 \cite{stable-baselines3}. The RL training of a neural network model is initialized with a ROS node through this framework.

    The RL learning agent is replaced with a ROS-based RL learning agent, also the main control node of a system, which interacts with the simulation environment as in the middle image in Fig  \ref{fig:first_figure} through sensor modules, implemented by \cite{pybullet_ros}, that publish data over ROS. The learning agent continuously receives sensory data at a rate from separately threaded sensor modules, different from standard RL where data is only updated in every timestep simulation of the simulator. 
    
\begin{figure}[tbp]
\centering
\includegraphics[width=0.47\textwidth]{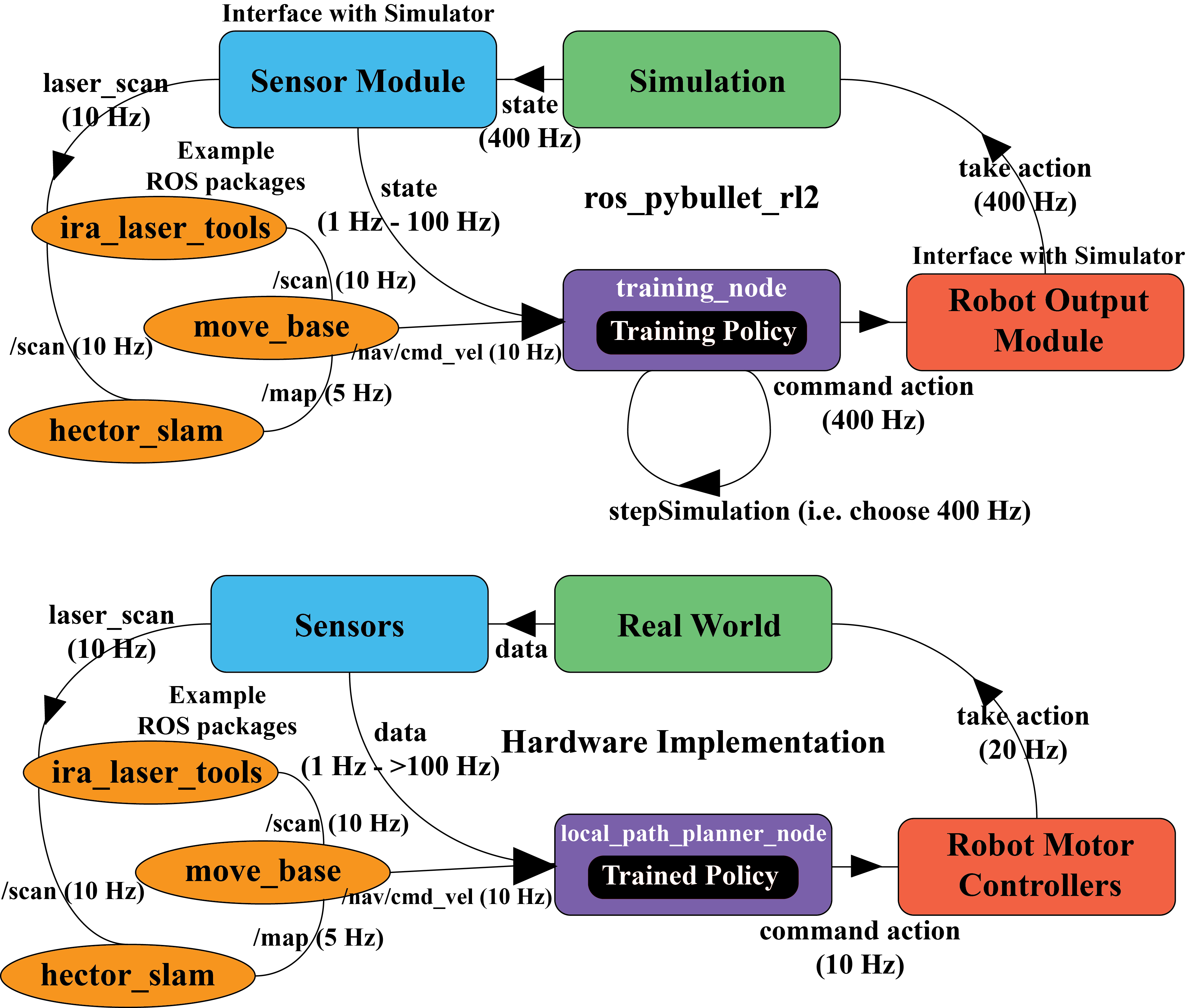}
\caption{A simplified ROS system architecture in operating a RL framework with ROS. The difference in the rate of interaction of the ROS topic information among nodes is observed between (Top) running RL training in simulation and (Bottom) running the trained policy in practice.}
\label{fig:sim_vs_real}
\vspace{-4mm}
\end{figure}

    The robot agent is able to run data preprocessing modules before passing data to the RL algorithm as inputs for learning. After further processing of the generated output action, the final velocity is published over ROS and received by a motor control module in the form of a plugin to execute the low-level trajectory command to take an action in the simulator.

    One advantage of \textit{ros\_pybullet\_rl2} is the interoperability with other ROS packages to avoid the need for reimplementation across different platforms. Fig.  \ref{fig:sim_vs_real} demonstrates a sample use of \textit{ros\_pybullet\_rl2} to connect RL with other ROS packages whether as a data processing step prior input into a neural network or as a node generating an output for motor control. When the RL environment initializes the ROS agent, it is similar to activating a robotic platform in real-life, but instead governing the entire robotic system interaction based on simulation physics. 

    The advantage of using the proposed framework is the integration of the ROS robotic middleware into the simulation to improve simulation-to-real transferability. 
    Through this integrated framework, given that mostly similar algorithms and pipelines used in simulation are also used in real-life application, the gap between the systems running in both dimensions are reduced. Secondly, the operating frequency of a ROS-based sensor and control output modules can be varied to match in both dimensions as well. Moreover, external ROS packages may be integrated with the simulated robotic system in RL training for a close-to-full-picture real world robotic system. The last remaining issue is in matching the simulation environmental data to the real world, and this discrepancy may be addressed through the use of sensor data processing methods and normalization techniques.

\subsection{Reinforcement Learning}

    An end-to-end RL procedure is designed to train an optimal policy that maps from minimally processed robot sensor data to the output trajectory to be executed by the robot. This trained policy will be used to form the base of a heuristic-based local path planning algorithm, with the benefits of largely reducing the reliance on preprocessed data, reducing computational load, giving rise to a more agile solution for obstacle avoidance in crowded environments. 

    A combined total of 32 normalized input observation states is fed into the neural network for learning. The network input state space is parameterized as follows:

\begin{equation}
    s=[d_{g},\theta_{g},v_{nav}, v_{prev}, F_{c}, d_{l}]
\end{equation}\label{eq:state_space}
\vspace{-2mm}

    \textit{Information on goal:} Firstly, to induce knowledge of the distance to the goal, $d_{g}$ is calculated by the Euclidean distance $d_{g} = \mid p_{goal}-p_{robot}\mid$. The learning agent should identify the direction of the goal using the heading angle to the goal in the robot frame. $\theta_g$ is reduced to the range [-180\textdegree, 180\textdegree]. 
    
    \textit{Velocity control:} The velocity command output from the EBand local planner in the current timestep, $v_{nav}$, is used as an input to the neural network as a recommended trajectory produced by the EBand local path planner. The command velocity taken by the robot in simulation in the previous timestep, $v_{prev}$, is also used as an input to the neural network. This allows the robot agent to take an action referencing its velocity in the previous timestep. 
    
    \textit{Collision query:} To introduce a tactile aware intelligence layer, the robot model is equipped with 6 contact sensor plates aimed at detecting collision queries from all directions. The contact sensor input, $[F_{c}]$, is calibrated to correspond with the real-life force sensor readings, with a maximum of 10 kg loading. The reading indicates the amount of load exerted on the robot by contact objects or obstacles.
    
    \textit{Obstacle proximity:} The laser scan data is a processed output from the package \textit{ira\_laser\_tool} which merges the laser scan data from two opposite-facing LiDARs into a full 360° field of view around the robot. The maximum detection range is 30 m. The laser measurements, $[d_{l}]$, are compressed into 18 values via minimum pooling and subjected to Gaussian noise with a standard deviation $\sigma = 0.01$.

    The output of the neural network features a continuous action space in 3 dimensions, x, y and $\theta$ velocities. The generated action produces a vector summation with the velocity output by the EBand local planner at equal weights ($w_{1}=w_{2}=1.0$) represented by the following: 
\begin{gather}
	\begin{bmatrix} 
    v_{x} \\
	v_{y} \\
	\omega_{z} \\
	\end{bmatrix}_{output}
	 = 
	w_{1} \times
	\begin{bmatrix} 
	v_{x} \\
	v_{y} \\
	\omega_{z} \\
	\end{bmatrix}_{Policy}
	+ 
	w_{2} \times
	\begin{bmatrix} 
	v_{x} \\
	v_{y} \\
	\omega_{z} \\
	\end{bmatrix}_{EBand}
	\label{eq:sum_cmd_vel}
\end{gather}
    
    The calculated velocity is clipped and scaled according to the maximum possible motor output speed in real-life. If the processed velocity falls below a certain threshold, it is considered deadzone velocity and reduced to zero. %

    The network architecture used is a Multi-Layer Perceptron (MLP) \cite{stable-baselines3} consisting of 4 hidden layers each with 64 nodes activated by a ReLU function, and an output layer that branches into a policy network and a value network. 
    Using the Proximal Policy Optimization (PPO) RL algorithm \cite{stable-baselines3}, a feedforward neural network inheriting features from both Advantage Actor Critic (A2C) and Trust Region Policy Optimization (TRPO) \cite{schulman2017trust} takes in a list of 32 input observations to train a policy to output an optimal action consisting of 3 dimensions. 

    The reward function is the metadata which enables the network to learn which actions are favourable given the observation inputs at every timestep. We formulate the reward policy as follows:
\begin{equation}
    r(s_{t})= \begin{cases} 10 & d_{g} < 0.3 \\
                               -0.05 & d_{g,t} - d_{g,t-1} > 0 \\
                               0.01 & d_{g,t} - d_{g,t-1} < 0 \\
                               a = -0.02 & \mid{\theta_{g,t}} - {\theta_{g,t-1}}\mid  \geqslant 0 \\
                               a = 0 & \theta_{g} \leqslant 0.5235 \\
                                a = -0.1\mid{\theta_{g}}\mid & \mid\theta_{g}\mid  > 0.5235 \\
                               - K_{p} \times (d_{p} - \min (d_{l_{1,.., 18}})) \\
                               - K_{f} \times \min (F_{c_{1,.., 6}})
                \end{cases} \label{eq:reward_fn}
\end{equation}

    where $d_{g}$ is the distance where a detectable object is in close proximity, $K_{p}$ and $K_{f}$ are scaling factor expressing weightage of collision and proximity, and $a$ is the cost due to the angle displacement from the target. $K_{p}$ and $K_{f}$ are determined through satisfactory performance in obstacle avoidance experiments using artificial potential field (APF) algorithm \cite{APF_khatib}. 
    
    The reward function awards the agent for reaching the goal and penalizes severely when colliding with dynamic obstacles or static obstacles according to $- K_{f} \times \min (F_{c_{1, 2, ..., 6}})$. In addition, close proximity to dynamic obstacles or objects is penalized slightly to prevent extreme close contact most of the time by $- K_{p} \times (d_{p} - \min (d_{l_{1,2, ..., 18}}))$ below a threshold distance $d_{p} = 0.5 \ m$. 
    The robot is subject to a higher penalty if it does not move towards the goal than not facing the goal so as to discourage non-progressive movements. In general, the design of the reward policy is to keep the mean reward range per episode small within [-10,10] where convergence can be expected around 0. The final goal of the agent is to choose actions at each time step in order to maximize its expected future discounted reward, given by:
    
    \begin{equation}
       \max E \sum_{t=0}^{\infty} \gamma^{t} R(s_t,a_t) \label{eq}
    \end{equation}
   
    In every training episode, a \textit{hector\_slam} node and \textit{move\_base} node are started to do live simultaneous localization and mapping (SLAM) as well as both global and local path planning. The neural network learns to output a trajectory in every timestep to reach the goal. 
    Our RL model is trained to reach progressive goals with increasing complexity (i.e. increasing distance and turning, goals behind static walls and around corridors, and in environments with dynamic obstacles moving randomly within a range of velocities [-0.5, 0.5] $m/s$).

    One training environment is initialized at a time. The RL training is run using the PPO algorithm \cite{schulman2017proximal, stable-baselines3} for 1 million timesteps, where $dt=0.01\ s$ in simulation, with 10 epochs and 32 minibatches. 5 policy evaluation episodes are run every 10,000 training timesteps. 

    The training hyperparameters were optimized base on experiments in our simulation environment and presented here. The learning rate is chosen at $1\times 10^{-4}$,  discount factor $\gamma = 0.99$, entropy coefficient $\beta = 1\times 10^{-2}$, clipping range $\epsilon$ = 0.2, generalized advantage estimator \cite{schulman2018highdimensional_gae} $\lambda$ = 0.95 and Adam optimizer \cite{kingma2017adam} is used. An episode is terminated when the goal is reached, collision is experienced in 100 timesteps within an episode, or after 1,000 timesteps elapsed. 

\section{Experiments}

\subsection{Computational Details}

Our framework, \textit{ros\_pybullet\_rl2}, has been tested on and runs on ROS Kinetic 16.04 and Melodic 18.04 distributions. Instructions have been released for both setups with the dependencies required described in the documentation of the open-source repository. %
The RL model was trained on a computer with Intel Core i7-4710HQ CPU with 8 cores. Hardware testing was done on the robot's onboard computer which is an Intel Core i7-7500U Kabylake CPU with 4 cores. 

Offline training takes approximately 8 hours to complete 500,000 timesteps. A ROS implementation of the trained policy as a local path planning algorithm is %
run on the robot at 10 Hz, which matches the average rate a normal human takes to react to environmental stimulation \cite{VANRULLEN2012995_10Hz}. 

\subsection{Simulation Results}

The \textit{ros\_pybullet\_rl2} trained neural network model is incorporated as a trajectory optimizer for the existing A* global path planning combined with EBand local path planner method, an implementation of the ROS Navigation Stack. For evaluation of the model, we present two test cases in the following sections to validate the performance of the RL-based local path planning algorithm for goal-reaching and obstacle avoidance behaviour.

For validation in the training simulation environment, our proposed solution, A*-EBand-RL, is compared to a combined A*-EBand-APF \cite{apf}, GA3C-CADRL \cite{everett2018motion_cadrl}, A*-EBand \cite{291936eband}, and A*-DWA \cite{dwa}. For the state-of-the-art GA3C-CADRL, the trained policy provided by the author is used. A*-EBand and A*-DWA are existing classical path planning approaches by ROS Navigation Stack. In all test cases, an onmidirectional robot is set in an unknown environment with no pre-loaded static map and no prior information. SLAM is done by a \textit{hector\_slam} node, and the live map is used by \textit{move\_base} node for path planning.

\begin{figure}[tbp]
\centering
\subfloat[A*-EBand-RL]{\includegraphics[width= 1in]{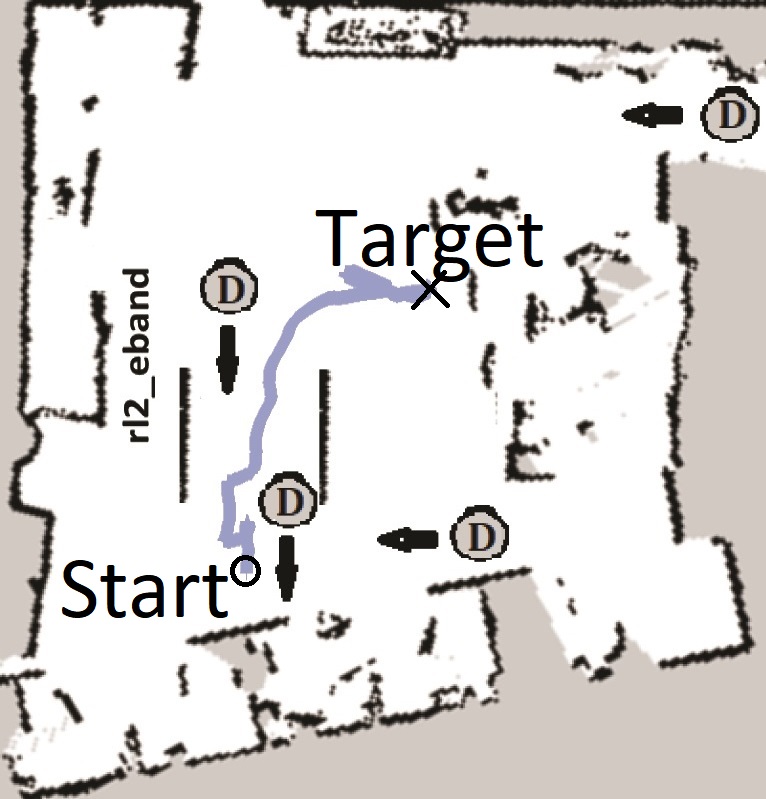}}
\hspace{1pt}
\subfloat[CADRL]{\includegraphics[width= 1in]{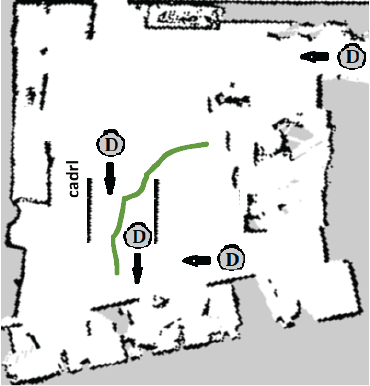}}
\hspace{1pt}
\subfloat[A*-EBand-APF]{\includegraphics[width= 1in]{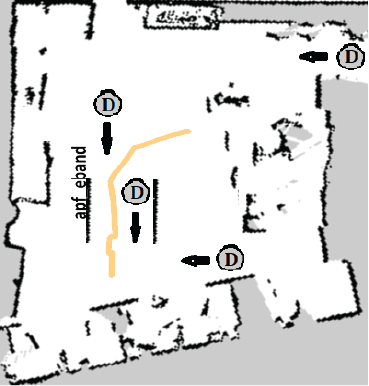}} \\
\subfloat[A*-EBand]{\includegraphics[width= 1in]{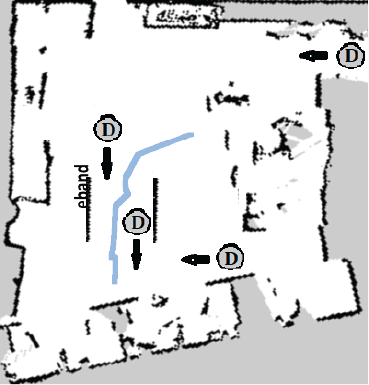}}
\hspace{1pt}
\subfloat[A*-DWA]{\includegraphics[width= 1in]{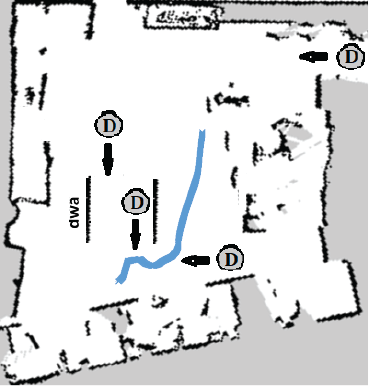}}
\caption{Overlay of graphs showing the path taken by all algorithms in one instance of the \textit{lab} environment. Circles marked with "D" represent the dynamic obstacles.}
\label{fig:trained_sim_env}
\vspace{-4mm}
\end{figure}

Tests were conducted in a simulation environment named \textit{lab (Dynamic)}, on which the RL model has been trained on before. This test environment is selected as it is easy for the robot to localize itself in a relatively cluttered place with many identifiable features. The robot is instructed to travel to the same final goal position among static obstacles, with a few dynamic obstacles moving at random in the simulation environment. The predefined direction of motions of the dynamic obstacles can be seen in Fig. \ref{fig:trained_sim_env}. 

\begin{figure}[tbp]
\centering
\includegraphics[width=0.4\textwidth]{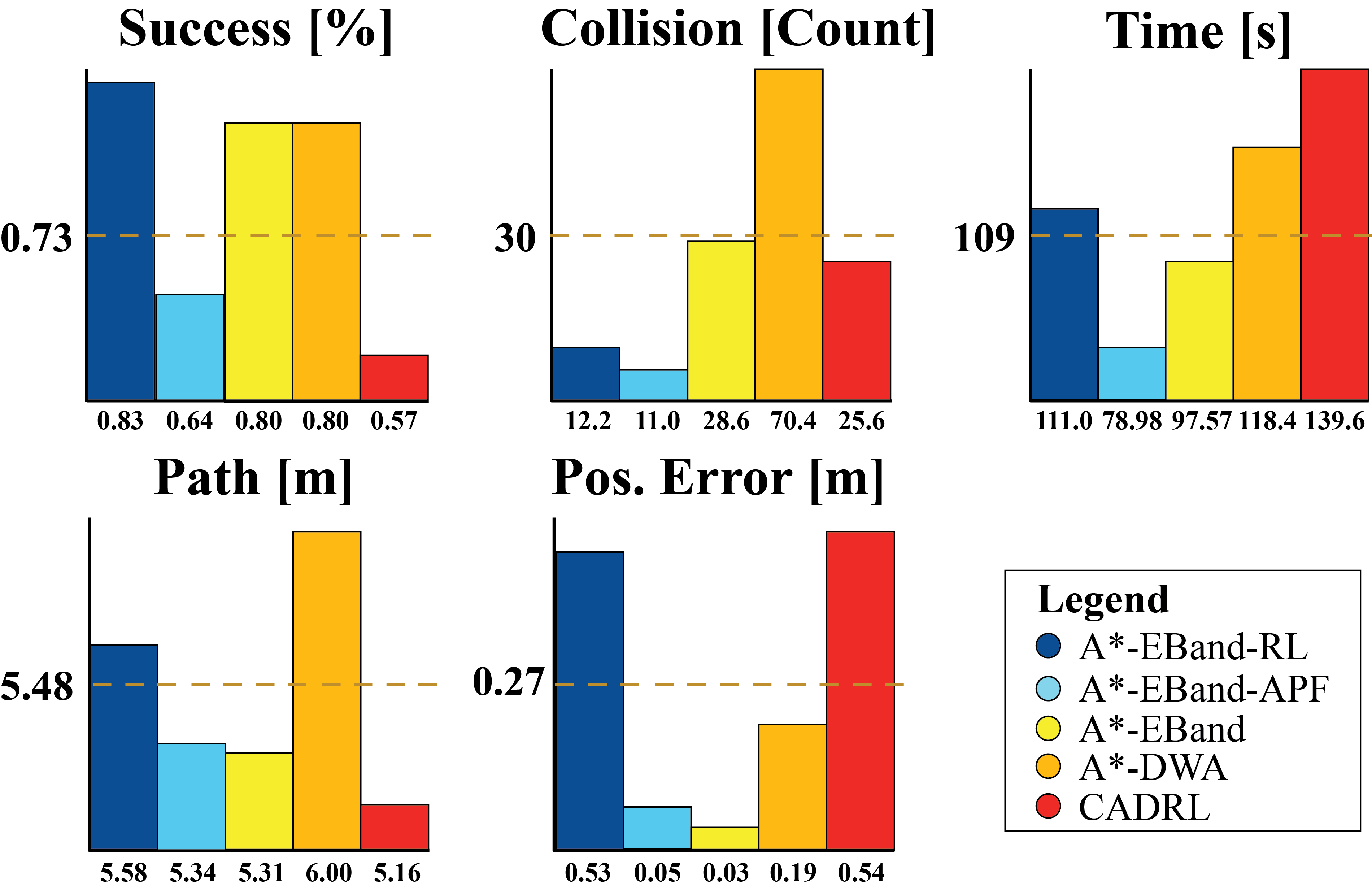}
\vspace{-2mm}
\caption{A bar chart showing the metrics comparison of the different navigation methods in trained unknown simulation environments. The labels represent the actual values achieved by each method, and the dotted line shows the average performance.}
\label{fig:bar_chart}
\vspace{-3mm}
\end{figure}

We compare the performance of the different algorithms using the metrics: success rate [$\%$], instances of collision, average time taken to reach goal, average path length [$m$] and final position error [$m$] (See Fig.  \ref{fig:bar_chart}). Success rate describes the fraction of cases where the robot agent reaches the goal position without the algorithm failing. The instances of collision is tabulated as 1 in every timestep regardless of whether one or more contact sensors register collision queries in that timestep. Time taken shows how long it took for the robot to reach the goal and average path length indicates how far the robot travelled to reach the goal. Position error is measured as the difference between the robot's final position and the goal.

 The dataset from 6 trials of each algorithm were curated to give the results in Fig. \ref{fig:bar_chart}. All algorithms are found to have average speeds ranging from 0.27 to 0.31. From the results, it can be seen that our proposed method, A*-EBand-RL has the highest success rate, second lowest number of collisions, and performed adequately in terms of path length, position error, and time taken. The path taken by A*-EBand-RL in Fig. \ref{fig:trained_sim_env} when it avoided the incoming obstacle shows it moving away and backwards, and from this we can see that the trained model has learned to move back in view of approaching obstacles from the front. This behavior would also help to explain why the average path length and position error is higher than some of the classical path planning algorithms. 
 
 CADRL presents short path lengths with smooth trajectories, but will experience collision if it does not happen to detect dynamic obstacles coming in its way in unseen unknown environments. CADRL do not have much success in this test scenario as the algorithm do not work well in unknown environments without accurate waypoints to take that will allow it to avoid running into static obstacle. 
 
 A*-EBand-APF presents good performance in the number of collisions and time with low error in position. The APF is traditionally known to be easily trapped in a local minima especially when more than two obstacles come around the robot agent or when it requires to pass through a tight corridor. The low success rate may be explained by the repulsive force of APF which causes perturbed actions in the proximity of obstacles, leading the robot to deviate largely from its path, which results in the EBand planner experiencing robot pose acquisition problem. Methods like A*-EBand and A*-DWA seems to have a high success rate but experiences especially high volume of collisions because the $move\_base$ planner is not sufficiently reactive, requiring more time to be able to replan especially in dynamic environments with constantly changing costmap, taking a long time to replan its path. A*-EBand-RL on the other hand, overcomes this limitation as the RL velocity trajectory optimizer will not fail to run even when the EBand planner fails in calculating a path and trajectory command, allowing unperturbed navigation.

\begin{figure}[tbp]
\centering
\includegraphics[width=0.48\textwidth]{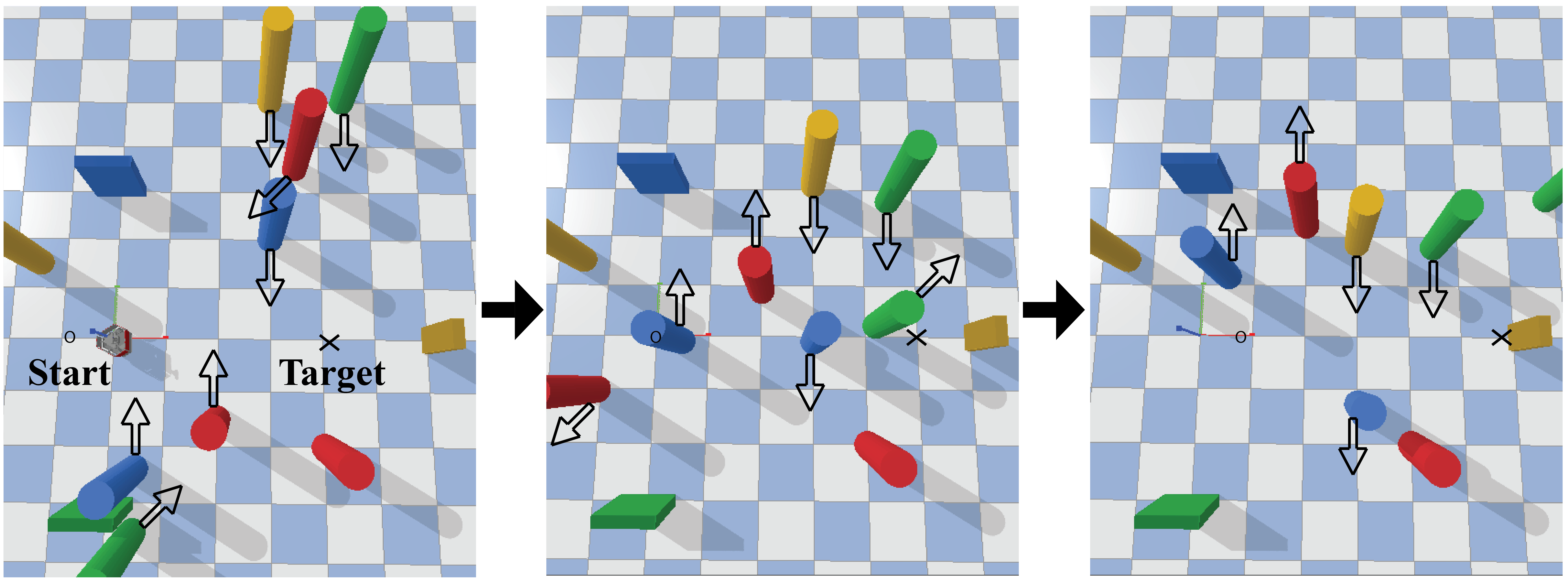}
\vspace{-5mm}
\caption{A snippet of the simulation showing the changes in the simulation environment due to motions taken by dynamic obstacles.}
\label{fig:untrained_sim}

\vspace{3mm}

\includegraphics[width=0.48\textwidth]{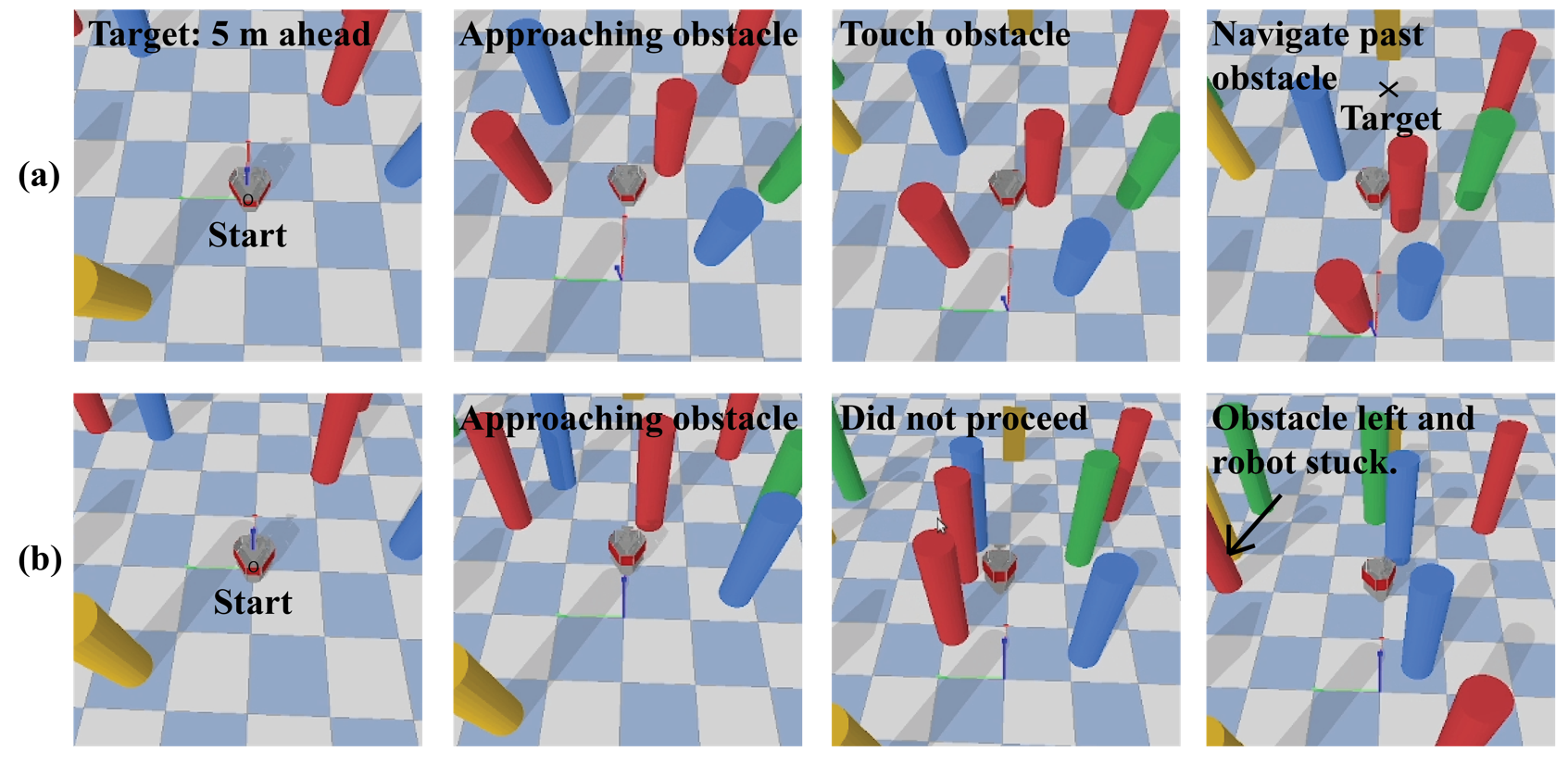}
\vspace{-5mm}
\caption{The simulation snippets showing the difference in performance between (a) A*-EBand-RL and (b) A*-EBand.}
\label{fig:rlvs}
\vspace{-3mm}
\end{figure}

Further tests were conducted in an untrained simulation environment named \textit{open\_ground (Dynamic)}. %
This test environment is selected as it is difficult for the robot to localize itself in an open ground with many moving objects. The dynamic obstacles move in a fixed motion in every run of the simulation setup shown in Fig. \ref{fig:untrained_sim}. In this scenario, A*-EBand-RL is compared against CADRL only. The other algorithms have been omitted from Table \ref{tab1e:metrics_2} because the classical algorithms had zero successes in reaching the target. Here we added a new metric for comparison, the root-mean-square (RMS) error [$m$] which describes the extent of deviation in the path taken by the different algorithms relative to a ground truth path - illustration of the ideal path to navigate to the target. Six trials of each algorithm were tested in this environment and we obtain the results in Table \ref{tab1e:metrics_2}. The A*-EBand-RL surpasses CADRL in most performance metrics. 
Despite the existence of the costmap generated by the \textit{move\_base} node, A*-EBand-RL is able to go up close to obstacles to move around obstacles, with slight collisions at times. Through these simulation results, we find that A*-EBand-RL is able to generalise to unseen, crowded dynamic environments with acceptable performance.

\begin{table}[tbp]
\caption{Metrics Comparison of the RL-based navigation methods in untrained unknown simulation environments.}
\vspace{-3mm}
\begin{center}
\begin{tabular}{c|c c c c c}
Algorithm & Success & Collision & Time & Path & RMSE \\
\hline
A*-EBand-RL & \textbf{0.83} & \textbf{8.2} & \textbf{16.5} & 14.6 & \textbf{2.31} \\
CADRL & 0.66 & 258.6 & 55.6 & \textbf{13.2} & 3.05 \\
\hline
Improvement & 25.7\% & 96.8\% & 70.3\% & (10.6\%) & 24.3\% \\
\end{tabular}
\label{tab1e:metrics_2}
\end{center}
\vspace{-6mm}
\end{table}

To test the efficacy of the framework and to validate the use of a RL-trained tactile aware intelligence layer for local path planning, we implement the application on a real-life mobile robot. 

\subsection{Hardware Experiment}

The 3-wheeled omnidirectional mobile robot in Fig. \ref{fig:robot} is positioned in an indoor environment largely similar to that of the simulation environment with many static obstacles and 3 to 5 dynamic obstacles. The robot is instructed to travel to the same goal position as defined in simulation runs. The robot performance using A*-EBand-RL largely resembles that of the simulation - able to avoid incoming dynamic obstacles, does not show confused behaviour when brushing against moving obstacles on opposite sides of the robot, and does not move too close to static obstacles. When compared to the A*-EBand-APF, the RL-based method is superior in navigating doorways through intelligent behaviour such as giving way or moving around dynamic obstacles without algorithmic failure. Overall, the hardware experiment shows successful simulation-to-real transferability of the trained policy from the \textit{ros\_pybullet\_rl2} framework. 

A hardware demonstration video can be found at \href{https://bit.ly/ros-pybullet-rl}{https://bit.ly/ros-pybullet-rl}.
\section*{Conclusion}

This work presents: 1. a tactile layer for collision detection queries to complement the merged LiDAR data for dynamic obstacle avoidance in a crowded environment on an omnidirectional mobile robot platform, 2. a RL framework integrating ROS, Pybullet simulation, and OpenAI Gym, 3. an end-to-end trained policy that generates %
trajectories for dynamic obstacle avoidance, and 4. an RL-based local path planning algorithm with a high runtime efficiency %
for implementation on robotic hardware. In simulation environments, the trained neural network model has shown great success in navigating the various scenarios with efficient paths and few to little collisions. Our real-world experiments has shown success in navigating among dynamic obstacles such as pedestrians in real life, %
and our approach could be further exposed to a greater variety of training environments to improve the ability to generalise.

Future works are ongoing for the addition of processed RGB-D camera data for training of a neural network for dynamic obstacle avoidance and human intention recognition. Works on making the RL framework support parallel ROS training environment through the use of containers are also ongoing. 

The proposed \textit{ros\_pybullet\_rl2} framework is made open-source at: \href{https://github.com/mcx-lab/ros\_pybullet\_rl2}{https://github.com/mcx-lab/ros\_pybullet\_rl2}.

\balance
\bibliographystyle{unsrt}
\bibliography{root.bbl}

\end{document}